\documentclass[runningheads]{llncs}
\usepackage{graphicx}
\usepackage{amsmath,amssymb} 
\usepackage{color}
\usepackage{multicol}
\usepackage{algorithm}
\usepackage{subfigure}
\usepackage{algorithmic}
\usepackage{multirow}
\usepackage[width=122mm,left=12mm,paperwidth=146mm,height=193mm,top=12mm,paperheight=217mm]{geometry}
\usepackage{subfigure}

\begin{document}

\mainmatter
\def\ECCVSubNumber{100}  

\title{Gradient Centralization: A New Optimization Technique for Deep Neural Networks} 
\authorrunning{}
\titlerunning{Gradient Centralization}
\authorrunning{H. Yong et al.}


\author{Hongwei Yong\inst{1,2} \and
Jianqiang Huang\inst{2} \and
Xiansheng Hua\inst{2} \and
Lei Zhang\inst{1,2}}


\institute{Department of Computing, The Hong Kong Polytechnic University
\email{\{cshyong,cslzhang\}@comp.polyu.edu.hk}\\
 \and
DAMO Academy, Alibaba Group\\
\email{\{jianqiang.jqh,huaxiansheng\}@gmail.com}}
\maketitle

\begin{abstract}
Optimization techniques are of great importance to effectively and efficiently train a deep neural network (DNN). It has been shown that using the first and second order statistics (e.g., mean and variance) to perform Z-score standardization on network activations or weight vectors, such as batch normalization (BN) and weight standardization (WS), can improve the training performance. Different from these existing methods that mostly operate on activations or weights, we present a new optimization technique, namely gradient centralization (GC), which operates directly on gradients by centralizing the gradient vectors to have zero mean.  GC can be viewed as a projected gradient descent method with a constrained  loss function. We show that GC can  regularize both the weight space and output feature space so that it can  boost the generalization performance of DNNs.
Moreover, GC improves the Lipschitzness of the  loss function and its gradient so that the training process becomes more efficient and stable.  GC is very simple to implement and can be easily embedded into existing gradient based DNN optimizers with only one line of code. It can also be directly used to fine-tune the pre-trained DNNs. Our experiments on various applications, including general image classification, fine-grained image classification, detection and segmentation, demonstrate that GC can consistently improve the performance of DNN learning. The code of GC can be found at https://github.com/Yonghongwei/Gradient-Centralization.

\end{abstract}

\section{Introduction}
\vspace{-2mm}
The broad success of deep learning largely owes to the recent advances on large-scale datasets~\cite{russakovsky2015imagenet}, powerful computing resources (e.g., GPUs and TPUs), sophisticated network architectures~\cite{he2016deep,huang2017densely} and optimization algorithms~\cite{bottou1991stochastic,kingma2014adam}. Among these factors, the efficient optimization techniques, such as stochastic gradient descent (SGD) with momentum~\cite{qian1999momentum}, Adagrad ~\cite{duchi2011adaptive} and  Adam~\cite{kingma2014adam}, make it possible to train  very deep neural networks (DNNs) with a large-scale dataset, and consequently deliver more powerful and robust DNN models  in practice. The generalization performance of the trained DNN models as well as the efficiency of training process depend essentially on the employed optimization techniques.

There are two major goals for a good DNN optimizer: accelerating the training process and improving the model generalization capability. The first goal aims to spend less time and cost to reach a good local minima, while the second goal aims to ensure that the learned DNN model can make accurate predictions on test data. A variety of optimization algorithms~\cite{qian1999momentum,duchi2011adaptive,kingma2014adam,duchi2011adaptive,kingma2014adam} have been proposed to achieve these goals. SGD ~\cite{bottou1991stochastic,bottou2010large} and its extension SGD with momentum (SGDM)~\cite{qian1999momentum} are among the most commonly used ones. They update the parameters along the opposite direction of their gradients in one training step. Most of the current DNN optimization methods are based on SGD and improve SGD to better overcome the gradient vanishing or explosion problems. A few successful techniques have been proposed, such as weight initialization strategies~\cite{glorot2010understanding,he2015delving}, efficient active functions (e.g., ReLU~\cite{nair2010rectified}), gradient clipping~\cite{pascanu2012understanding,pascanu2013difficulty}, adaptive learning rate optimization algorithms~\cite{duchi2011adaptive,kingma2014adam}, and so on.

In addition to the above techniques, the sample/feature statistics such as mean and variance can also be used to normalize the network activations or weights to make the training process more stable. The representative methods operating on activations include batch normalization (BN)~\cite{ioffe2015batch}, instance normalization (IN)~\cite{ulyanov2016instance,huang2017arbitrary}, layer normalization (LN)~\cite{lei2016layer} and group normalization (GN)~\cite{wu2018group}. Among them, BN is the most widely used optimization technique which normalizes the features along the sample dimension in a mini-batch for training. BN smooths the optimization landscape~\cite{Santurkar2018How} and it can speed up the training process and boost model generalization performance when a proper batch size is used~\cite{Zhang2016Deep,he2016deep}. However, BN works not very well when the training batch size is small, which limits its applications to memory consuming tasks, such as object detection~\cite{he2017mask,ren2015faster}, video classification~\cite{karpathy2014large,abu2016youtube}, etc.

Another line of statistics based methods operate on weights. The representative ones include weight normalization (WN)~\cite{salimans2016weight,huang2017centered} and weight standardization (WS)~\cite{qiao2019weight}. These methods re-parameterize weights to restrict weight vectors during training. For example, WN decouples the length of weight vectors from their direction to accelerate the training of DNNs. WS uses the weight vectors' mean and variance to standardize them to have zero mean and unit variance. Similar to BN, WS can also smooth the loss landscape and speed up training. Nevertheless,  such methods operating on weight vectors cannot directly adopt the pre-trained models (e.g., on ImageNet) because their weights may not meet the condition of zero mean and unit variance.

\begin{figure}[t]
\centering
\subfigure[]
{\includegraphics[width=0.46\textwidth]{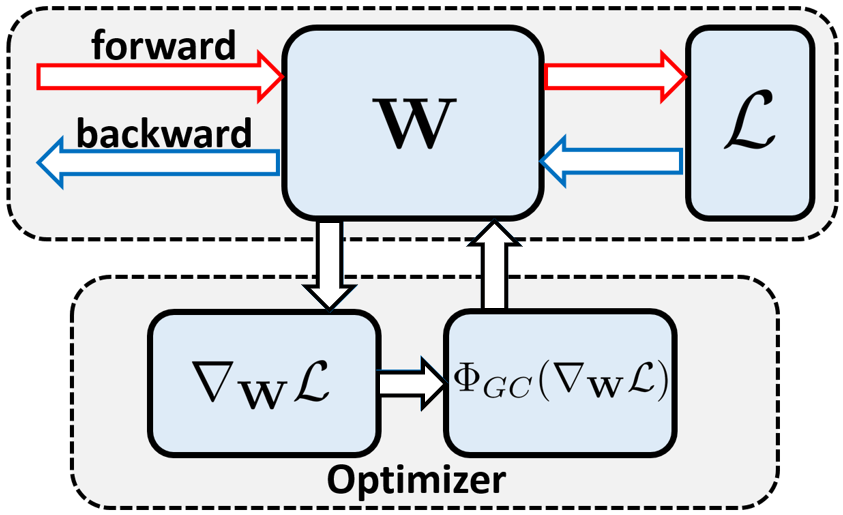}}
\subfigure[]
{\includegraphics[width=0.52\textwidth]{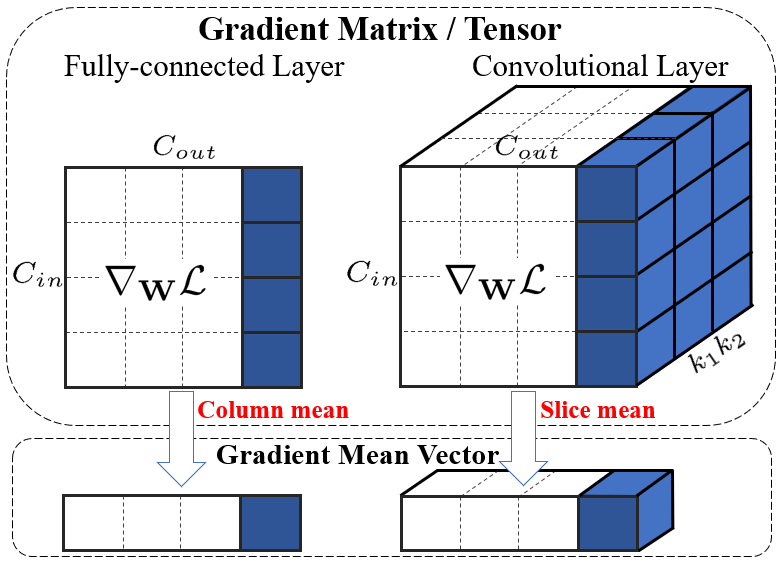}}
\vspace{-3mm}
\caption{(a) Sketch map for using gradient centralization (GC). $\mathbf{W}$ is the weight, $\mathcal{L}$ is the loss function, $\nabla_{\mathbf{W}}\mathcal{L}$ is the gradient of weight, and $\Phi_{GC}(\nabla_{\mathbf{W}}\mathcal{L})$ is the centralized gradient. It is very simple to embed GC into existing network optimizers by replacing $\nabla_{\mathbf{W}}\mathcal{L}$ with $\Phi_{GC}(\nabla_{\mathbf{W}}\mathcal{L})$.
(b) Illustration of the GC operation on gradient matrix/tensor of weights in the fully-connected layer (left) and convolutional layer (right). GC computes the column/slice mean of gradient matrix/tensor and centralizes each column/slice to have zero mean.
 }
\label{F:gradient}\vspace{-5mm}
\end{figure}

Different from the above techniques which operate on activations or weight vectors, we propose a very simple yet effective DNN optimization technique, namely gradient centralization (GC), which operates on the gradients of weight vectors. As illustrated in Fig. \ref{F:gradient}(a), GC simply centralizes the gradient vectors to have zero mean. It can be easily embedded into the current gradient based optimization algorithms (e.g., SGDM~\cite{qian1999momentum}, Adam~\cite{kingma2014adam}) using only one line of code. Though simple, GC demonstrates various desired properties, such as accelerating the training process, improving the generalization performance, and the compatibility for fine-tuning pre-trained models. The main contributions of this paper are highlighted as follows:
    \begin{itemize}
   \item[$\bullet$] We propose a new general network optimization technique, namely gradient centralization (GC), which can not only smooth and accelerate the training process of DNN but also improve the model generalization performance.
   \item[$\bullet$] We analyze the theoretical properties of GC, and show that GC constrains the loss function by introducing a new constraint on weight vector, which regularizes both the weight space and output feature space so that it can boost model generalization performance. Besides, the constrained loss function has better Lipschitzness than the original one, which makes the training process more stable and efficient.
   \end{itemize}
\vspace{-1mm}

Finally, we perform comprehensive experiments on various applications, including general image classification, fine-grained image classification, object detection and instance segmentation. The results demonstrate that GC can consistently improve the performance of learned DNN models in different applications. It is a simple, general and effective network optimization method.

\vspace{-2mm}

\section{Related Work}
\vspace{-2mm}

In order to accelerate the training and boost the generalization performance of DNNs, a variety of optimization techniques~\cite{ioffe2015batch,wu2018group,salimans2016weight,qiao2019weight,qian1999momentum,pascanu2012understanding} have been proposed to operate on activation, weight and gradient. In this section we briefly review the related work from these three aspects.

\textbf{Activation:}
The activation normalization layer has become a common setting in DNN, such as batch normalization (BN)~\cite{ioffe2015batch} and group normalization (GN)~\cite{wu2018group}. BN was originally introduced to solve the internal covariate shift by normalizing the activations along the sample dimension. It  allows higher learning rates~\cite{Bjorck2018Understanding}, accelerates the training speed and improves the generalization accuracy~\cite{luo2018towards,Santurkar2018How}. However, BN does not perform well when the training batch size is small, and GN is proposed to address this problem by normalizing the activations or feature maps in a divided group for each input sample. In addition, layer normalization (LN) ~\cite{lei2016layer} and instance normalization (IN)~\cite{ulyanov2016instance,huang2017arbitrary} have been proposed for RNN and style transfer learning, respectively.

\textbf{Weight:}
 Weight normalization (WN)~\cite{salimans2016weight} re-parameterizes the weight vectors and decouples the length of a weight vector from its direction. It speeds up the convergence of SGDM algorithm to a certain degree. Weight standardization (WS)~\cite{qiao2019weight} adopts the Z-score standardization to re-parameterize the weight vectors. Like BN, WS can also smooth the loss landscape and improve training speed. Besides, binarized DNN~\cite{rastegari2016xnor,courbariaux2016binarized,courbariaux2015binaryconnect} quantifies the weight into binary values, which can improve the generalization capability for certain DNNs. However, a shortcoming of those methods operating on weights is that they cannot be directly used to fine-tune pre-trained models since the pre-trained weight may not meet their constraints. As a consequence, we have to design specific pre-training methods for them  in order to fine-tune the model.

\textbf{Gradient:}
A commonly used operation on gradient is to compute the momentum of gradient~\cite{qian1999momentum}. By using the momentum of gradient, SGDM accelerates SGD in the relevant direction and dampens oscillations. Besides, $L_2$ regularization based weight decay, which introduces $L_2$ regularization into the gradient of weight, has long been a standard trick to improve the generalization performance of DNNs~\cite{krogh1992simple,zhang2018three}. To make DNN training more stable and avoid gradient explosion, gradient clipping~\cite{pascanu2012understanding,pascanu2013difficulty,abadi2016deep,kim2016accurate} has been proposed to train a very deep DNNs. In addition, the projected gradient methods~\cite{gupta2018cnn,larsson2017projected} and Riemannian approach~\cite{cho2017riemannian,vorontsov2017orthogonality} project the gradient on a subspace or a Riemannian manifold to regularize the learning of weights.

\vspace{-2mm}
\section{Gradient Centralization}
\vspace{-2mm}
\subsection{Motivation}

BN~\cite{ioffe2015batch} is a powerful DNN optimization technique, which uses the first and second order statistics to perform Z-score standardization on activations. It has been shown in ~\cite{Santurkar2018How} that BN reduces the Lipschitz constant of loss function and makes the gradients more Lipschitz smooth so that the optimization landscape  becomes smoother. WS~\cite{qiao2019weight} can also reduce the Lipschitzness of loss function and smooth the optimization landscape through Z-score standardization on weight vectors. BN and WS operate on activations and weight vectors, respectively, and they implicitly constrict the gradient of weights, which improves the Lipschitz property of loss for optimization.

Apart from operating on activation and weight, can we directly operate on gradient to make the training process more effective and stable? One intuitive idea is that we use Z-score standardization to normalize gradient, like what has been done by BN and WS on activation and weight. Unfortunately, we found that normalizing gradient cannot improve the stability of training. Instead, we propose to compute the mean of gradient vectors and centralize the gradients to have zero mean. As we will see in the following development, the so called gradient centralization (GC) method can have good Lipschitz property, smooth the DNN training and improve the model generalization performance.

\vspace{-2mm}
\subsection{Notations }
\vspace{-1mm}

We define some basic notations. For fully connected layers (FC layers), the weight matrix is denoted as
$\mathbf{W}_{fc} \in \mathbb{R}^{C_{in}\times C_{out}}$, and for  convolutional layers (Conv layers) the weight tensor is denoted as
$\mathbf{W}_{conv}  \in \mathbb{R}^{C_{in}\times C_{out}\times(k_1k_2)}$, where $C_{in}$ is the number of input channels, $C_{out}$ is the number of output channels, and $k_1$, $k_2$ are the kernel size of convolution layers. For the convenience of expression, we unfold the weight tensor of Conv layer into a matrix/tensor and use a unified notation $\mathbf{W} \in \mathbb{R}^{M\times N}$ for weight matrix in FC layer ($\mathbf{W}\in \mathbb{R}^{C_{in}\times C_{out}}$) and Conv layers ($\mathbf{W}\in  \mathbb{R}^{(C_{in}k_1k_2)\times C_{out}}$).
Denote by $\mathbf{w}_i\in\mathbb{R}^{M}$ ($i=1,2,...,N$) the $i$-th column vector of weight matrix $\mathbf{W}$ and
$\mathcal{L}$  the objective function.  $\nabla_{\mathbf{W}}\mathcal{L}$ and $\nabla_{\mathbf{w}_i}\mathcal{L}$
denote the gradient of $\mathcal{L}$ w.r.t. the weight matrix $\mathbf{W}$ and weight vector $\mathbf{w}_i$,
respectively. The size of gradient matrix $\nabla_{\mathbf{W}}\mathcal{L}$  is the same as weight matrix $\mathbf{W}$. Let  $\mathbf{X}$ be the input activations for this layer and $\mathbf{W}^T\mathbf{X}$ be its output activations. $\mathbf{e}=\frac{1}{\sqrt{M}}\mathbf{1}$ denotes an $M$ dimensional unit vector and $\mathbf{I}\in\mathbb{R}^{M\times M}$ denotes an identity matrix.

\vspace{-2mm}
\subsection{Formulation of GC}
\vspace{-1mm}
For a FC layer or a Conv layer, suppose that we have
obtained the gradient through  backward  propagation, then for a weight vector $\mathbf{w}_{i}$ whose gradient is $\nabla_{\mathbf{w}_{i}}\mathcal{L}$ ($i=1,2,...,N$), the GC operator, denoted by $\Phi_{GC}$, is defined as follows:
\vspace{-1mm}
\begin{equation}
\begin{aligned}
\Phi_{GC}(\nabla_{\mathbf{w}_{i}}\mathcal{L})=\nabla_{\mathbf{w}_{i}}\mathcal{L}-\mu_{\nabla_{\mathbf{w}_{i}}\mathcal{L}}
  \end{aligned}\label{GZ_vector}
  \vspace{-2mm}
\end{equation}
where $\mu_{\nabla_{\mathbf{w}_{i}}\mathcal{L}}=\frac{1}{M}\sum_{j=1}^M \nabla_{\mathbf{W}_{i,j}}\mathcal{L}$. The formulation of GC is very simple. As shown in Fig. \ref{F:gradient}(b), we only need to compute the mean of the column vectors of the weight matrix, and then remove the mean from each column vector.  We can also have a matrix formulation of Eq. (\ref{GZ_vector}):
\vspace{-1mm}
\begin{equation}
\begin{aligned}
\Phi_{GC}(\nabla_{\mathbf{W}}\mathcal{L})=\mathbf{P}\nabla_{\mathbf{W}}\mathcal{L}, \ \ \ \ \ \mathbf{P}=\mathbf{I}-\mathbf{e}\mathbf{e}^{T}
  \end{aligned}\label{GZ_matrix}
  \vspace{-1mm}
\end{equation}

The physical meaning of $\mathbf{P}$ will be explained later in Section \ref{Imp_Gen}. In practical implementation, we can directly remove the mean value from each weight vector to accomplish the GC operation. The computation is very simple and efficient.

\vspace{-2mm}
\subsection{Embedding of GC to SGDM/Adam}
\vspace{-1mm}
GC can be easily embedded into the current DNN optimization algorithms such as SGDM~\cite{qian1999momentum,bottou2010large} and  Adam~\cite{kingma2014adam}.
After obtaining the centralized gradient $\Phi_{GC}(\nabla_{\mathbf{w}}\mathcal{L})$, we can directly use it to update the weight matrix. Algorithm \ref{alg1} and Algorithm  \ref{alg2} show  how to embed GC into the two most popular optimization algorithms, SGDM and Adam, respectively. Moreover, if we want to use weight decay, we can set $\mathbf{\widehat{g}}^t=\mathbf{P}(\mathbf{g}^t+\lambda\mathbf{w})$, where $\lambda$ is the weight decay factor. It only needs to add one line of code into most existing DNN optimization algorithms to execute GC with negligible additional computational cost. For example, it costs only  $0.6$ sec extra training time in one epoch on CIFAR100 with ResNet50 model in our experiments  (71 sec for one epoch).

\vspace{-5mm}
\begin{algorithm}
\caption{SGDM with Gradient Centralization}
\begin{multicols}{2}
\begin{algorithmic}[1]\small
\renewcommand{\algorithmicrequire}{\textbf{Input:}}
\renewcommand{\algorithmicensure}{\textbf{End}}
\REQUIRE  Weight vector $\mathbf{w}^0$, step size $\alpha$,  momentum factor $\beta$, $\mathbf{m}^0$
\renewcommand{\algorithmicrequire}{\textbf{Training step:}}
\renewcommand{\algorithmicensure}{\textbf{End}}
\REQUIRE
\FOR{$t=1,...T$}
\STATE
$\mathbf{g}^t=\nabla_{\mathbf{w}^t}\mathcal{L}$
\STATE
$\mathbf{\widehat{g}}^t=\Phi_{GC}(\mathbf{g}^t)$
\STATE
$\mathbf{m}^t=\beta\mathbf{m}^{t-1}+(1-\beta)\mathbf{\widehat{g}}^t$
\STATE
$\mathbf{w}^{t+1}=\mathbf{w}^t-\alpha\mathbf{m}_t$
\ENDFOR
\end{algorithmic}
\end{multicols}
\vspace{-3mm}\label{alg1}
\end{algorithm}

\vspace{-14mm}
\begin{algorithm}
\caption{Adam with Gradient Centralization}
\begin{multicols}{2}
\begin{algorithmic}[1]\small
\renewcommand{\algorithmicrequire}{\textbf{Input:}}
\renewcommand{\algorithmicensure}{\textbf{End}}
\REQUIRE Weight vector $\mathbf{w}^0$, step size $\alpha$,  $\beta_1$, $\beta_2$, $\epsilon$, $\mathbf{m}^0$,$\mathbf{v}^0$
\renewcommand{\algorithmicrequire}{\textbf{Training step:}}
\renewcommand{\algorithmicensure}{\textbf{End}}
\REQUIRE
\FOR{$t=1,...T$}
\STATE
$\mathbf{g}^t=\nabla_{\mathbf{w}^t}\mathcal{L}$
\STATE
$\mathbf{\widehat{g}}^t=\Phi_{GC}(\mathbf{g}^t)$
\STATE
$\mathbf{m}^t=\beta_1\mathbf{m}^{t-1}+(1-\beta_1)\mathbf{\widehat{g}}^t$
\STATE
$\mathbf{v}^t=\beta_2\mathbf{v}^{t-1}+(1-\beta_2)\mathbf{\widehat{g}}^t\odot\mathbf{\widehat{g}}^t$
\STATE
$\mathbf{\widehat{m}}^t=\mathbf{m}^t/(1-(\beta_1)^t)$
\STATE
$\mathbf{\widehat{v}}^t=\mathbf{v}^t/(1-(\beta_2)^t)$
\STATE
$\mathbf{w}^{t+1}=\mathbf{w}^t-\alpha\frac{\mathbf{\widehat{m}}^t}{\sqrt{\mathbf{\widehat{v}}^t}+\epsilon}$
\ENDFOR
\end{algorithmic}
\end{multicols}
\vspace{-4mm}\label{alg2}
\end{algorithm}

\vspace{-6mm}
\section{Properties of GC}
\vspace{-2mm}
As we will see in the section of experimental result, GC can accelerate the training process and improve the generalization performance of DNNs. In this section, we perform theoretical analysis to explain why GC works.

\vspace{-3mm}
\subsection{Improving Generalization Performance}\label{Imp_Gen}

\vspace{-1mm}
One important advantage of GC is that it can improve the generalization performance of DNNs. We explain this advantage from two aspects: weight space regularization and output feature space regularization.

\vspace{1mm}
\textbf{Weight space regularization:}
 Let's first explain the physical meaning of $\mathbf{P}$ in Eq.(\ref{GZ_matrix}). Actually, it is easy to prove that:
\vspace{-2mm}
\begin{equation}
\begin{aligned}
\mathbf{P}^2=\mathbf{P}=\mathbf{P}^T, \ \ \ \ \ \ \mathbf{e}^T\mathbf{P}\nabla_{\mathbf{W}}\mathcal{L}=0.
 \end{aligned}
 \vspace{-2mm}
\end{equation}
The above equations show that $\mathbf{P}$ is the projection matrix for the hyperplane with normal vector $\mathbf{e}$ in weight space, and $\mathbf{P}\nabla_{\mathbf{W}}\mathcal{L}$ is the projected
gradient.

\begin{figure}[t]
\centering
\includegraphics[width=0.65\textwidth]{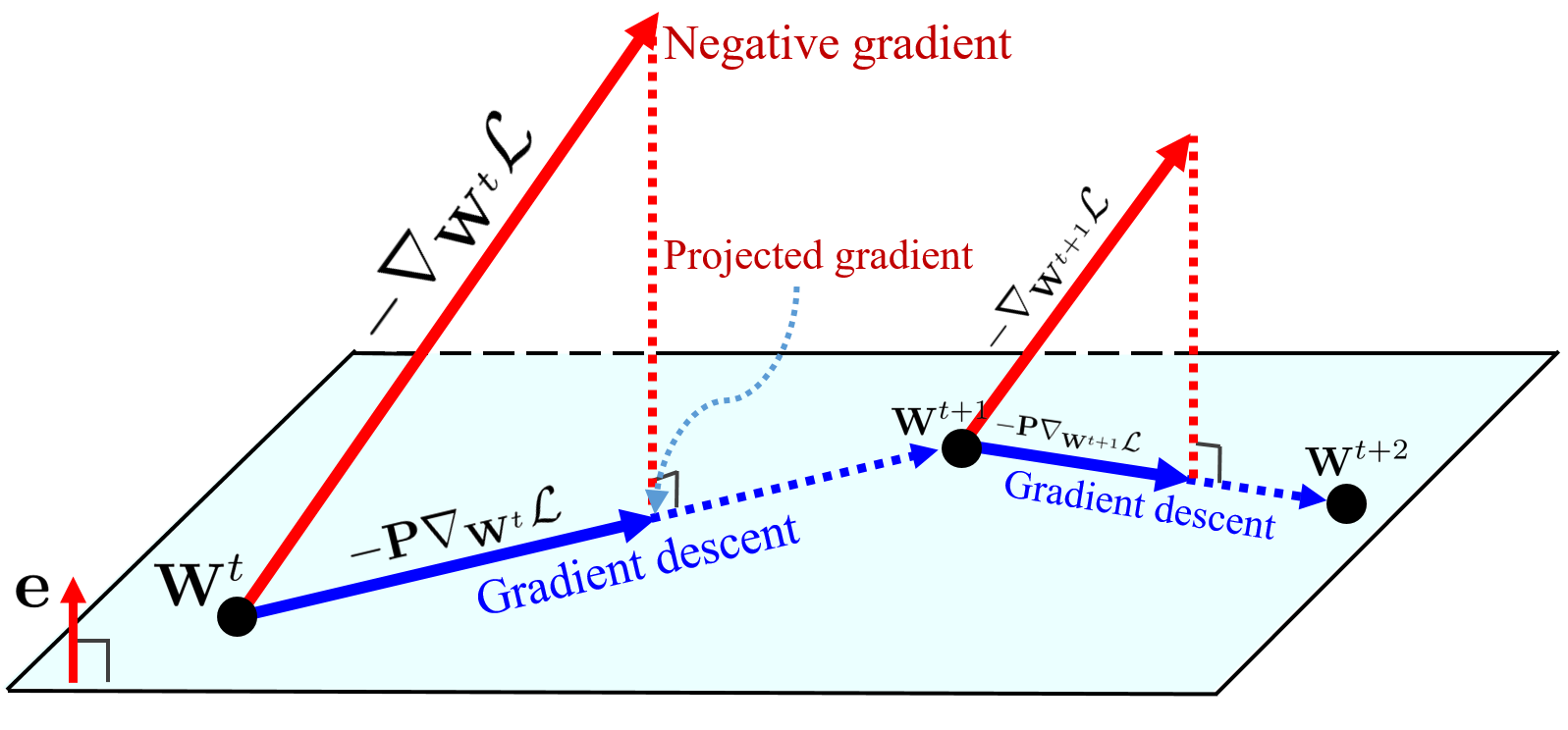}\vspace{-5mm}
\caption{The geometrical interpretation of GC. The gradient is projected on a hyperplane  $\mathbf{e}^T(\mathbf{w}-\mathbf{w}^t)=0$, where the projected gradient is used to update the weight.}
\label{F:Project}\vspace{-6mm}
\end{figure}

The property of projected gradient has been investigated in some previous works~\cite{gupta2018cnn,larsson2017projected,cho2017riemannian,vorontsov2017orthogonality}, which indicate that projecting the gradient of weight will constrict the weight space in a hyperplane or a Riemannian manifold. Similarly, the role of GC can also be viewed from the perspective of projected gradient descant.
We give a geometric illustration of SGD with GC in Fig. \ref{F:Project}. As shown in Fig. \ref{F:Project},
 in the $t$-th step of SGD with GC, the gradient is first projected on the hyperplane determined by $\mathbf{e}^T(\mathbf{w}-\mathbf{w}^t)=0$, where $\mathbf{w}^t$ is the weight vector in the $t$-th iteration, and then the weight is updated along the direction of projected gradient $-\mathbf{P}\nabla_{\mathbf{w}^t}\mathcal{L}$. From  $\mathbf{e}^T(\mathbf{w}-\mathbf{w}^t)=0$, we have $\mathbf{e}^T\mathbf{w}^{t+1}=\mathbf{e}^T\mathbf{w}^t=...=\mathbf{e}^T\mathbf{w}^0$, i.e., $\mathbf{e}^T\mathbf{w}$ is a constant during training.
 Mathematically, the latent objective function w.r.t. one weight vector $\mathbf{w}$ can be written as follows:
 \vspace{-1mm}
\begin{equation}
\begin{aligned}
\min_{\mathbf{w}}\mathcal{L}(\mathbf{w}), \ \ \ \ s.t. \ \ \ \ \mathbf{e}^T(\mathbf{w}-\mathbf{w}^0)=0
  \end{aligned}\label{latent_fun}
  \vspace{-3mm}
\end{equation}
Clearly, this is a constrained optimization problem on weight vector $\mathbf{w}$.
It regularizes the solution space of $\mathbf{w}$, reducing the possibility of over-fitting on training data.
As a result, GC can improve the generalization capability of trained DNN models, especially when the number of training samples is limited.

It is noted that WS~\cite{qiao2019weight}  uses a constraint $\mathbf{e}^T\mathbf{w}=0$ for weight optimization. It  reparameterizes weights to meet this constraint. However, this constraint largely
limits its practical applications because the initialized weight may not satisfy this constraint. For example, a pretrained DNN on ImageNet usually cannot meet $\mathbf{e}^T\mathbf{w}^0=0$ for its initialized weight vectors. If we use WS to fine-tune this DNN, the advantages of pretrained models will disappear. Therefore, we have to retrain the DNN on ImageNet with WS before we fine-tune it. This is very cumbersome. Fortunately the weight constraint of GC in Eq. (\ref{latent_fun}) fits any initialization of weight, e.g., ImageNet pretrained initialization, because
 it involves the initialized weight $\mathbf{w}^0$ into the constraint so that
 $\mathbf{e}^T(\mathbf{w}^0-\mathbf{w}^0)=0$ is always true.
This  greatly extends the applications of GC.

\vspace{1mm}
\textbf{Output feature space regularization:}
For SGD based algorithms, we have
$\mathbf{w}^{t+1}=\mathbf{w}^{t}-\alpha^t\mathbf{P}\nabla_{\mathbf{w}^t}\mathcal{L}$. It can be derived that $\mathbf{w}^{t}=\mathbf{w}^{0}-\mathbf{P}\sum_{i=0}^{t-1}\alpha^{(i)}\nabla_{\mathbf{w}^{(i)}}\mathcal{L}$.
 For any input feature vector $\mathbf{x}$, we have the following theorem:
\vspace{1mm}

\noindent
\textbf{Theorem 4.1:}\label{Theorem2}
\emph{Suppose that SGD (or SGDM) with GC  is used to update the weight vector $\mathbf{w}$, for any input feature vectors $\mathbf{x}$ and $\mathbf{x}+\gamma\mathbf{1}$, we have
\vspace{-2mm}
\begin{equation}
\begin{aligned}
({\mathbf{w}^{t}})^T\mathbf{x}-({\mathbf{w}^{t}})^T(\mathbf{x}+\gamma\mathbf{1})=\gamma\mathbf{1}^T\mathbf{w}^0
\end{aligned}
\vspace{-2mm}
\end{equation}
where $\mathbf{w}^0$ is the initial weight vector and $\gamma$ is a scalar.
\vspace{1mm}
}

Please find the proof in the \textbf{Appendix}. Theorem \ref{Theorem2} indicates that a constant intensity change (i.e., $\gamma\mathbf{1}$)  of an input feature causes a change of output activation; interestingly, this change is only related to $\gamma$ and $\mathbf{1}^T\mathbf{w}^0$ but not the current weight vector $\mathbf{w}^t$. $\mathbf{1}^T\mathbf{w}^0$  is the scaled mean of the initial weight vector $\mathbf{w}^0$. In particular, if the mean of $\mathbf{w}^0$  is close to zero, then the output activation is not sensitive to the intensity change of input features, and the output feature space becomes  more robust to training sample variations.

\begin{figure}[t]
\centering
\includegraphics[width=0.57\textwidth]{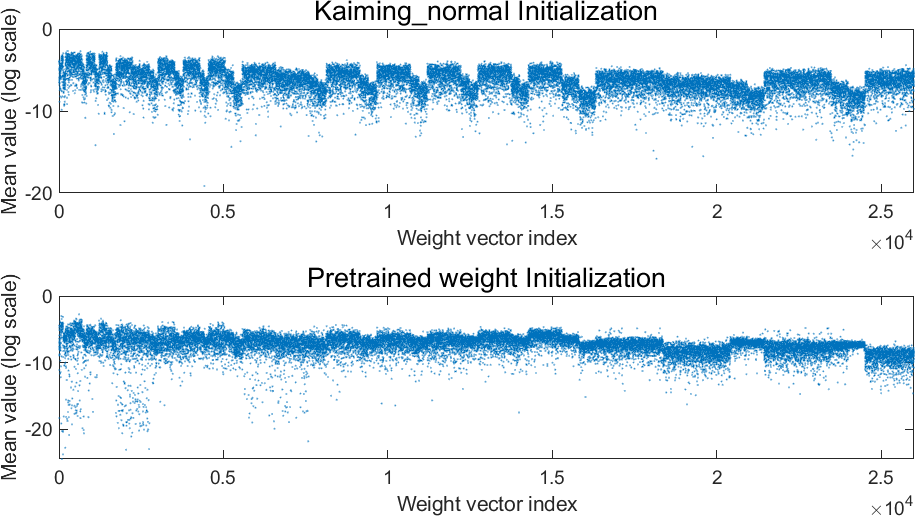}\vspace{-3mm}
\caption{The absolute value (log scale) of the mean of weight vectors for convolution layers in ResNet50. The $x$-axis is the weight vector index. We plot the mean value of different convolution layers from left to right with the order from sallow to deep layers. Kaiming normal initialization~\cite{he2015delving} (top) and ImageNet pre-trained weight initialization (bottom) are employed here. We can see that the mean values are usually very small (less than $e^{-7}$) for most of the weight vectors. }
\label{F:weight_init}\vspace{-6mm}
\end{figure}

Indeed, the mean of $\mathbf{w}^0$ is very close to zero by the commonly used weight initialization strategies, such as Xavier initialization~\cite{glorot2010understanding}, Kaiming initialization~\cite{he2015delving} and even ImageNet pre-trained weight initialization. Fig. \ref{F:weight_init} shows the absolute value (log scale) of the mean of weight vectors for Conv layers in ResNet50 with Kaiming normal initialization and ImageNet pre-trained weight initialization. We can see that the mean values of most weight vectors are very small and close to zero (less than $e^{-7}$). This ensures that if we train the DNN model with GC, the output features will not be sensitive to the variation of the intensity of input features. This property regularizes the output feature space and boosts the generalization performance of DNN training.

\vspace{-3mm}
\subsection{Accelerating Training Process}
\vspace{-1mm}
\noindent
\textbf{Optimization landscape smoothing:}
It has been shown in ~\cite{Santurkar2018How,qiao2019weight} that both BN and WS smooth the optimization landscape. Although BN and WS operate on activations and weights, they  implicitly constrict the gradient of weights, making the gradient of weight more predictive and stable for fast training. Specifically, BN and WS use the gradient  magnitude $||\nabla f(\mathbf{x})||_2$ to capture the Lipschitzness of function $f(\mathbf{x})$. For the loss and its gradients, $f(\mathbf{x})$ will be $\mathcal{L}$ and $\nabla_{\mathbf{w}}\mathcal{L}$, respectively,  and $\mathbf{x}$ will be $\mathbf{w}$. The upper bounds of $||\nabla_{\mathbf{w}}\mathcal{L}||_2$ and $||\nabla^2_{\mathbf{w}}\mathcal{L}||_2$ ($\nabla^2_{\mathbf{w}}\mathcal{L}$ is the Hessian matrix of $\mathbf{w}$)  have been given in~\cite{Santurkar2018How,qiao2019weight} to illustrate the optimization landscape smoothing property of BN and WS. Similar conclusion can be made for our proposed GC
by comparing the Lipschitzness of original loss function $\mathcal{L}(\mathbf{w})$ with the constrained loss function in Eq. (\ref{latent_fun}) and the Lipschitzness of their gradients.
We have the following theorem:

\noindent
\textbf{Theorem 4.2:}\label{Theorem1}
\emph{Suppose $\nabla_{\mathbf{w}}\mathcal{L}$ is the gradient of loss
 function $\mathcal{L}$ w.r.t. weight vector $\mathbf{w}$. With the $\Phi_{GC}(\nabla_{\mathbf{w}}\mathcal{L})$ defined in Eq.(\ref{GZ_matrix}), we have the following conclusion for the loss function and its gradient, respectively:
 \vspace{-2mm}
 \begin{equation}
 \left\{
\begin{aligned}
&||\Phi_{GC}(\nabla_{\mathbf{w}}\mathcal{L})||_2\leq||\nabla_{\mathbf{w}}\mathcal{L}||_2,\\
&||\nabla_{\mathbf{w}}\Phi_{GC}(\nabla_{\mathbf{w}}\mathcal{L})||_2\leq||\nabla^2_{\mathbf{w}}\mathcal{L}||_2.
\end{aligned}
\right.
\vspace{-4mm}
\end{equation}
}

The proof of Theorem \ref{Theorem1} can be found in the \textbf{Appendix}.
Theorem \ref{Theorem1} shows that for the loss function $\mathcal{L}$ and its gradient
$\nabla_{\mathbf{w}}\mathcal{L}$,
the constrained loss function in Eq. (\ref{latent_fun}) by GC leads to a better Lipschitzness
than the original loss function so that the optimization landscape becomes smoother. This means that GC has similar advantages to BN and WS on accelerating training. A good Lipschitzness on gradient implies that the gradients used in training are more predictive and well-behaved so that the optimization landscape can be smoother for faster and more effective training.

\begin{figure}[t]
\centering
\includegraphics[width=1\textwidth]{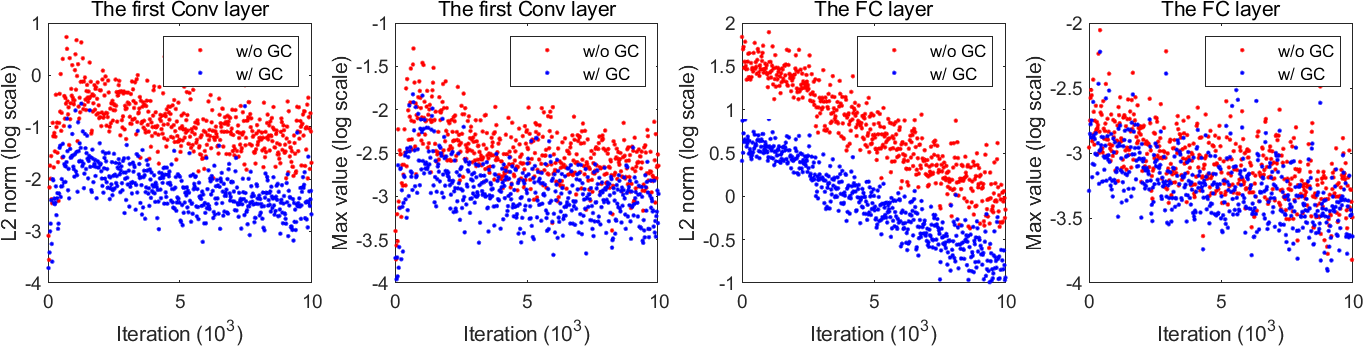}\vspace{-3mm}
\caption{The $L_2$ norm (log scale) and max value (log scale) of gradient matrix or tensor vs. iterations.  ResNet50 trained on CIFAR100 is used as the DNN model here. The left two sub-figures show the results on the first Conv layer and the right two show the FC layer. The red points represent the results of training without GC and the blue points represent the results with GC. We can see that GC largely reduces the $L_2$ norm and max value of gradient. }
\label{F_grad_clip}\vspace{-6mm}
\end{figure}

\vspace{1mm}
\textbf{Gradient explosion suppression:}
Another benefit of GC for DNN training is that GC can avoid gradient explosion and make training more stable. This property is similar to gradient clipping~\cite{pascanu2012understanding,pascanu2013difficulty,kim2016accurate,abadi2016deep}. Too large gradients will make the weights change abruptly during training so that the loss may severely oscillate and hard to converge. It has been shown that gradient clipping can suppress large gradient so that the training can be more stable and faster~\cite{pascanu2012understanding,pascanu2013difficulty}. There are two popular gradient clipping approaches: element-wise value clipping~\cite{pascanu2012understanding,kim2016accurate} and norm clipping~\cite{pascanu2013difficulty,abadi2016deep}, which apply thresholding to element-wise value and gradient norm to gradient matrix, respectively. In order to investigate the influence of GC on clipping gradient, in Fig. \ref{F_grad_clip} we plot the max value and $L_2$ norm of gradient matrix of the first convolutional layer and the fully-connected layer in ResNet50 (trained on CIFAR100) with and without GC. It can be seen that both the max value and the $L_2$ norm of the gradient matrix become smaller by using GC in training. This is in accordance to our conclusion in Theorem \ref{Theorem1} that GC can make training process smoother and faster.

\vspace{-2mm}
\section{Experimental Results}
\vspace{-1mm}
\subsection{Setup of Experiments}
\vspace{-1mm}
Extensive experiments are performed to validate the effectiveness of GC. To make the results as comprehensive and clear as possible, we arrange the experiments as follows:
\vspace{-3mm}
    \begin{itemize}
   \item[$\bullet$] We start from experiments on the Mini-ImageNet dataset~\cite{vinyals2016matching} to
   demonstrate that GC can accelerate the DNN training process and improve the model generalization performance. We also evaluate the combinations of GC with BN and WS to show that GC can improve them for DNN optimization.
   \item[$\bullet$] We then  use  the CIFAR100 dataset~\cite{krizhevsky2009learning} to evaluate GC with various DNN optimizers (e.g., SGDM, Adam, Adagrad),  various DNN architectures (e.g., ResNet, DenseNet, VGG), and and different hyper-parameters.
   \item[$\bullet$]We then perform experiments on ImageNet~\cite{russakovsky2015imagenet} to demonstrate that GC also works well on large scale image classification, and show that GC can also work well with normalization methods other than BN, such as GN.
   \item[$\bullet$]We consequently perform experiments on four fine-grained image classification datasets (FGVC Aircraft~\cite{maji2013fine}, Stanford Cars~\cite{krause20133d}, Stanford Dogs~\cite{khosla2011novel} and CUB-200-2011~\cite{wah2011caltech}) to show that GC can be directly adopted to fine-tune the pre-trained DNN models and improve them.
   \item[$\bullet$]At last, we perform experiments on the COCO dataset~\cite{lin2014microsoft} to show that GC also works well on other tasks such as objection detection and segmentation.
   \end{itemize}
\vspace{0mm}

GC can be applied to either Conv layer or FC layer, or both of them. In all of our following experiments, if not specified, we always apply GC to both Conv and FC layers. Except for Section \ref{exp_cifar} where we embed GC into different DNN optimizers for test, in all other sections we embed GC into SGDM for experiments, and the momentum is set to 0.9. All  experiments are conducted under the Pytorch 1.3 framework and the GPUs are NVIDIA Tesla P100.

We would like to stress that no additional hyper-parameter is introduced in our GC method. Only one line of code is needed to embed GC into the existing optimizers,  while keeping all other settings remain unchanged. We compare the performances of DNN models trained with and without GC to validate the effectiveness of GC.
\begin{figure}[t]
\centering
\includegraphics[width=0.72\textwidth]{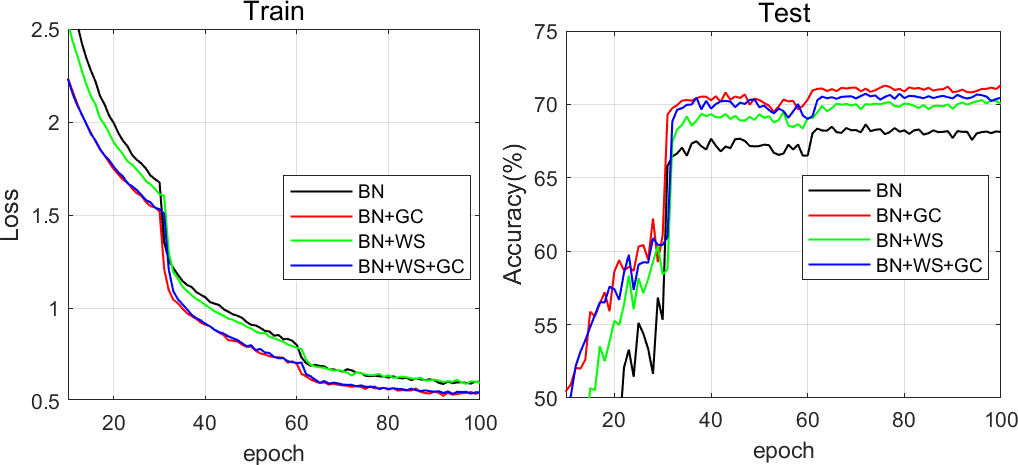}\vspace{-2mm}
\caption{Training loss (left) and testing accuracy (right) curves vs. training epoch on the Mini-ImageNet. The ResNet50 is used as the DNN model. The compared optimization techniques include BN, BN+GC, BN+WS and BN+WS+GC.}
\label{F_miniimagenet}\vspace{-5mm}
\end{figure}
\vspace{-2mm}
\subsection{Results on Mini-Imagenet}
\vspace{-1mm}

Mini-ImageNet~\cite{vinyals2016matching} is a subset of the ImageNet dataset~\cite{russakovsky2015imagenet}, which was originally proposed for few shot learning. We use the train/test splits provided by~\cite{ravi2016optimization,iscen2019label}. It consists of 100 classes and each class has 500 images for training and 100 images for testing. The image resolution is $84\times 84$. We resize the images into $224\times 224$, which is the standard ImageNet training input size. The DNN we used here is ResNet50, which is trained on 4 GPUs with batch size 128. Other settings are the same as training ImageNet. We repeat the experiments for 10 times and report the average results over the 10 runs.

BN, WS and GC operate on activations, weights and gradients, respectively, and they can be used together to train DNNs. Actually, it is necessary to normalize the feature space by methods such as BN; otherwise, the model is hard to be well trained. Therefore, we evaluate four combinations here: BN, BN+GC, BN+WS and BN+WS+GC. The optimizer is SGDM with momentum 0.9. Fig. $\ref{F_miniimagenet}$ presents the training loss and testing accuracy curves of these four combinations. Compared with BN, the training loss of BN+GC decreases much faster and the testing accuracy increases more rapidly.  For both BN and BN+WS, GC can further speed up their training speed.
Moreover, we can see that BN+GC achieves the highest testing accuracy, validating that GC can accelerate training and enhance the generalization performance simultaneously.
\vspace{-2mm}

\vspace{-0mm}
\subsection{Experiments on CIFAR100}\label{exp_cifar}
\vspace{-0mm}
\begin{table}[t]\scriptsize
\centering
\caption{Testing accuracies of different DNN models on CIFAR100}
\vspace{-2mm}
\begin{tabular}{c | c c c c c}%
\hline
Model &R18&R101&X29&V11&D121\\
\hline
w/o GC&76.87$\pm$0.26&78.82$\pm$ 0.42&79.70$\pm$0.30&70.94$\pm$ 0.34&79.31$\pm$0.33	\\
\hline
w/ GC&\textbf{78.82$\pm$0.31}	&\textbf{80.21$\pm$0.31}&	\textbf{80.53$\pm$0.33}&\textbf{71.69$\pm$0.37}&\textbf{79.68$\pm$0.40}\\
\hline
\end{tabular}\label{T:dnn_models}
\label{table1}
\vspace{0mm}
\end{table}

\begin{table}[t]\scriptsize
\centering
\caption{Testing accuracies of different optimizers on CIFAR100}
\vspace{-4mm}
\begin{tabular}[t]{c | c c c c c}
\hline
Algorithm &SGDM &Adam&Adagrad&SGDW&AdamW\\
\hline
w/o GC&78.23$\pm$0.42	&71.64$\pm$0.56&	70.34 $\pm$0.31& 74.02$\pm$0.27 &74.12$\pm$0.42	\\
\hline
w/ GC&\textbf{79.14$\pm$0.33}	&\textbf{72.80$\pm$0.62}&	\textbf{71.58$\pm$0.37}& \textbf{76.82$\pm$0.29} &\textbf{75.07$\pm$0.37}\\
\hline
\end{tabular}\label{table1}
\vspace{-4mm}
\end{table}

CIFAR100~\cite{krizhevsky2009learning} consists of 50K training images and 10K testing images from 100 classes. The size of input image is $32\times32$. Since the image resolution is small, we found that applying GC to the Conv layer is good enough on this dataset. All DNN models are trained for 200 epochs using one GPU with batch size 128. The experiments are repeated for 10 times and the results are reported in mean $\pm$ std format.
\vspace{-0mm}


\textbf{Different networks:}
We  testify GC on different DNN architectures, including ResNet18 (R18), ResNet101 (R101)~\cite{he2016deep}, ResNeXt29 4x64d (X29)~\cite{xie2017aggregated}, VGG11 (V11)~\cite{simonyan2014very}  and DenseNet121 (D121)~\cite{huang2017densely}. SGDM is used as the network optimizer. The weight decay is set to 0.0005. The initial learning rate is $0.1$ and it is multiplied by $0.1$ for every 60 epochs. Table \ref{T:dnn_models} shows the testing accuracies of these DNNs. It can be seen that the performance of all DNNs is improved by GC, which verifies that GC is a general optimization technique for different DNN architectures.
\vspace{-0mm}

\textbf{Different optimizers:}
We embed GC into different DNN optimizers, including SGDM~\cite{qian1999momentum}, Adagrad~\cite{duchi2011adaptive}, Adam~\cite{kingma2014adam}, SGDW and AdamW ~\cite{loshchilov2017decoupled}, to test their performance. SGDW and AdamW optimizers directly apply weight decay on weight, instead of using $L_2$ weight decay regularization. Weight decay is set to $0.001$ for SGDW and AdamW, and $0.0005$ for other optimizers.
 The initial learning rate is set to $0.1$, $0.01$ and $0.001$  for SGDM/SGDW, Adagrad, Adam/AdamW, respectively, and the learning rate is multiplied by $0.1$ for every 60 epochs. The other hyper-parameters are set by their default settings on Pytorch. The DNN model used here is ResNet50. Table \ref{table1} shows the testing accuracies. It can be seen that GC boosts the generalization performance of all the five optimizers. It is also found that adaptive learning rate based algorithms Adagrad and Adam have poor generalization performance on CIFAR100, while GC can improve their performance by $>0.9\%$.
\vspace{-0mm}

\textbf{Different hyper-parameter settings:} In order to illustrate that GC can achieve  consistent improvement with different hyper-parameters,
 we present the results of GC with different settings of weight decay and learning rates on the CIFAR100 dataset. ResNet50 is used as the backbone.
Table \ref{T:dnn_weightdecay} shows the testing accuracies with different settings of weight decay, including 0, $1e^{-4}$, $2e^{-4}$, $5e^{-4}$ and $1e^{-3}$. The optimizer is SGDM with learning rate $0.1$.    It can be seen that the performance of  weight decay  is consistently improved by GC.
Table \ref{T:dnn_Lr} shows the testing accuracies with different learning rates for SGDM and Adam. For SGDM, the learning rates are  $0.05$, $0.1$ and $0.2$, and for Adam, the learning rates are $0.0005$, $0.001$ and $0.0015$. The weight decay is set to  $5e^{-4}$. Other settings are the same as those in the manuscript. We can see that GC consistently improves the performance.

\begin{table}[t]\scriptsize
\centering
\caption{Testing accuracies of different weight decay on CIFAR100 with ResNet50.}
\vspace{-1mm}
\begin{tabular}{c | c c c c c c}%
\hline
Weight decay &0&$1e^{-4}$&$2e^{-4}$&$5e^{-4}$&$1e^{-3}$\\
\hline
w/o GC&71.62$\pm$0.31&73.91$\pm$0.35&75.57$\pm$0.33&78.23$\pm$0.42&77.43$\pm$0.30\\
\hline
w/ GC&\textbf{72.83$\pm$0.29}&\textbf{76.56$\pm$0.31}&\textbf{77.62$\pm$0.37}&\textbf{79.14$\pm$0.33}&\textbf{78.10$\pm$0.36}\\
\hline
\end{tabular}\label{T:dnn_weightdecay}
\vspace{-1mm}
\end{table}

\begin{table}[t]\scriptsize
\centering
\caption{Testing accuracies of different learning rates on CIFAR100 with ResNet50 for SGDM and Adam.}
\vspace{-1mm}
\begin{tabular}{c | c c c|c c c }%
\hline
Algorithm &SGDM&SGDM&SGDM&Adam&Adam&Adam\\
\hline
Learning rate &0.05&0.1&0.2&0.0005&0.001&0.0015\\
\hline
w/o GC&76.81$\pm$0.27&78.23$\pm$0.42&76.53$\pm$0.32&73.88$\pm$0.46&71.64$\pm$0.56&\textbf{70.63$\pm$0.44}\\
\hline
w/ GC&\textbf{78.12$\pm$0.33}&\textbf{79.14$\pm$0.33}&\textbf{77.71$\pm$0.35}&\textbf{74.32$\pm$0.55}&\textbf{72.80$\pm$0.62}&\textbf{71.22$\pm$0.49}\\
\hline
\end{tabular}\label{T:dnn_Lr}
\vspace{-6mm}
\end{table}

\begin{table}[t]\scriptsize
\centering
\caption{Top-1 error rates on ImageNet w/o GC and w/ GC.}
\vspace{-2mm}
\begin{tabular}{c | c c c c c}%
\hline
Datesets &R50BN&R50GN&R101BN&R101GN\\

\hline
w/o GC&  23.71 &  24.50  &  22.37   & 23.34\\
\hline
w/ GC&\textbf{23.21}  &  \textbf{23.53}  & \textbf{21.82} & \textbf{22.14}\\
\hline
\end{tabular}\label{table1_imagenet}
\vspace{-2mm}
\end{table}

\begin{figure}[t]
\centering
\includegraphics[width=0.75\textwidth]{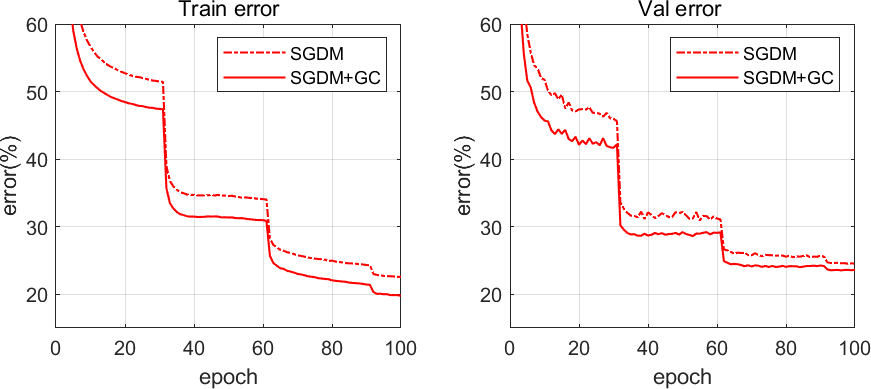}\vspace{-2mm}
\caption{Training error (left) and validation error (right) curves vs. training epoch on ImageNet. The DNN model is ResNet50 with GN.}
\label{F_imagenet}\vspace{-3mm}
\end{figure}

%
\vspace{-0mm}
\subsection{Results on ImageNet}
\vspace{-0mm}
We then evaluate GC on the large-scale ImageNet dataset~\cite{russakovsky2015imagenet} which consists of $1.28$ million images for training and 50K images for validation from 1000 categories. We use the common training settings and embed GC to SGDM on Conv layer. The ResNet50 and ResNet101 are used as the backbone networks. For the former, we use 4 GPUs with batch size 64 per GPU, and for the latter, we use 8 GPUs with batch size 32 per GPU.

We evaluate four models here: ResNet50 with BN (R50BN), ResNet50 with GN (R50GN), ResNet101 with BN (R101BN) and ResNet101 with GN (R101GN). Table \ref{table1_imagenet} shows the final Top-1 errors of these four DNN models trained with and without GC. We can see that GC can improve the performance by $0.5\%\sim 1.2\%$ on ImageNet. Fig. \ref{F_imagenet} plots the training and validation error curves of ResNet50 (GN is used for feature normalization). We can see that GC can largely speed up the training with GN.

\vspace{-3mm}

\begin{table}[t]\scriptsize
\centering
\caption{The statistics of fine-grained datasets used in this paper.}
\vspace{-2mm}
\begin{tabular}{c | c c c}%
\hline
Datasets&\#Category&\#Training&\#Testing\\
\hline
FGVC Aircraft&100&6,667&3,333\\
Stanford Cars&196&8,144&8,041\\
Stanford Dogs&120&12,000&8,580\\
CUB-200-2011&200&5,994&5,794\\
\hline
\end{tabular}\label{T_Fine_grained}
\vspace{-2mm}
\end{table}

\begin{table}[t]\scriptsize
\centering
\caption{Testing accuracies on the four fine-grained image classification datasets.}
\vspace{-2mm}
\begin{tabular}{c | c c c c c}%
\hline
Datesets &FGVC Aircraft&Stanford Cars&Stanford Dogs&CUB-200-2011\\
\hline
w/o GC &86.62$\pm$0.31&88.66$\pm$0.22 &76.16$\pm$0.25&82.07$\pm$0.26\\
w/ GC &\textbf{87.77$\pm$0.27}& \textbf{90.03$\pm$0.26}& \textbf{78.23$\pm$0.24}&\textbf{83.40$\pm$0.30}\\
\hline
\end{tabular}\label{T_fine-grained2}
\vspace{-6mm}
\end{table}

\subsection{Results on Fine-grained Image Classification}

In order to show that GC can also work with the pre-trained models, we conduct experiments on four challenging fine-grained image classification datasets, including FGVC Aircraft~\cite{maji2013fine}, Stanford Cars~\cite{krause20133d}, Stanford Dogs~\cite{khosla2011novel} and CUB-200-2011~\cite{wah2011caltech}. The detailed statistics of these four datasets are summarized in Table \ref{T_Fine_grained}. We use the official pre-trained ResNet50 provided by Pytorch as the baseline DNN for all these four datasets. The original images are resized into $512\times512$  and we crop the center region with $448\times448$ as input for both training and testing. The models are pre-trained on ImageNet. We use SGDM with momentum of $0.9$ to fine-tune ResNet50 for 100 epochs on 4 GPUs with batch size 256. The initial learning rate is $0.1$ for the last FC layer and 0.01 for all pre-trained Conv layers. The learning rate is multiplied by $0.1$ at the 50th and 80th epochs. Please note that our goal is to validate the effectiveness of GC but not to push state-of-the-art results, so we only use simple training tricks.  We repeat the experiments for 10 times and  report the result in mean $\pm$ std format.
\vspace{-6mm}

\begin{figure}
\centering
\includegraphics[width=1\textwidth]{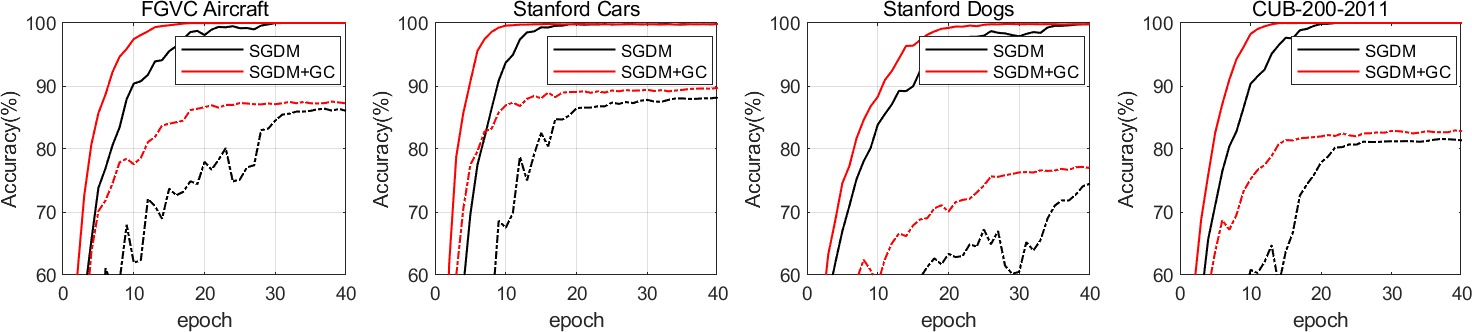}\vspace{-2mm}
\caption{Training accuracy (solid line) and testing accuracy (dotted line) curves vs. training epoch on four fine-grained image classification datasets.  }
\label{fine_grid2_c}\vspace{-7mm}
\end{figure}

Fig. \ref{fine_grid2_c} shows the training and testing accuracies of SGDM and SGDM+GC for the first 40 epochs on the four fine-grained image classification datasets. Table \ref{T_fine-grained2} shows the final testing accuracies. We can see that both the training and testing accuracies of SGDM are improved by GC. For the final classification accuracy, GC improves SGDM by $1.1\%\sim2.1\%$ on these four datasets. This sufficiently demonstrates the effectiveness of GC on fine-tuning pre-trained models.

\begin{table}[t]\scriptsize
\centering
\caption{Detection results on COCO by using Faster-RCNN and FPN with various backbone models.}
\vspace{-2mm}
\begin{tabular}{c| c |c c c||c |c c c }
\hline
Method & Backbone &$\text{AP}$& $\text{AP}_{.5}$ & $\text{AP}_{.75}$& Backbone &$\text{AP}$& $\text{AP}_{.5}$ & $\text{AP}_{.75}$  \\
\hline
w/o GC     &   R50       &36.4 &58.4 &39.1  &   X101-32x4d   &40.1 &62.0 &43.8  \\
w/ GC     &    R50       & 37.0   & 59.0   &  40.2  &    X101-32x4d    & 40.7   &  62.7  & 43.9 \\

\hline
w/o GC     &   R101      &38.5 &60.3 &41.6   &   X101-64x4d     & 41.3 &63.3& 45.2 \\
w/ GC     &    R101       & 38.9   & 60.8   & 42.2   &    X101-64x4d       & 41.6   & 63.8   & 45.4  \\
\hline
\end{tabular}\label{Tfaster}
\vspace{-1mm}
\end{table}

\begin{table}[t]\tiny
\centering
\caption{Detection and segmentation results on COCO by using Mask-RCNN and FPN with various backbone models.}
\vspace{-2mm}
\begin{tabular}{c| c |c c c |c c c || c |c c c |c c c}
\hline
Method & Backbone &$\text{AP}^b$& $\text{AP}^b_{.5}$ & $\text{AP}^b_{.75}$  & $\text{AP}^m$& $\text{AP}^m_{.5}$ & $\text{AP}^m_{.75}$ & Backbone &$\text{AP}^b$& $\text{AP}^b_{.5}$ & $\text{AP}^b_{.75}$  & $\text{AP}^m$& $\text{AP}^m_{.5}$ & $\text{AP}^m_{.75}$ \\
\hline
w/o GC     &   R50       &   37.4 &  59.0  &  40.6  & 34.1   & 55.5   &  36.1     &   R50 (4c1f)    &  37.5  &  58.2  &  41.0  &  33.9  &  55.0  &  36.1   \\
w/ GC     &    R50       & 37.9   & 59.6   & 41.2   & 34.7   & 56.1   &  37.0      &    R50 (4c1f)    &  38.4  & 59.5   &  41.8  & 34.6   & 55.9   & 36.7   \\

\hline
w/o GC     &   R101       &  39.4  &  60.9  & 43.3   &  35.9  & 57.7   &  38.4      &   R101GN        & 41.1   & 61.7   & 44.9   & 36.9   &  58.7  & 39.3  \\
w/ GC     &    R101       &  40.0  & 61.5   &  43.7  &  36.2  & 58.1   & 38.7      &    R101GN     & 41.7  &  62.3  &  45.3  &  37.4  & 59.3   &  40.3    \\
\hline
w/o GC     &   X101-32x4d      &  41.1& 62.8 &45.0 & 37.1 &59.4 &39.8       &   R50GN+WS       & 40.0   & 60.7   & 43.6   & 36.1   &  57.8  &  38.6\\
w/ GC     &    X101-32x4d     & 41.6  &  63.1 &  45.5  &  37.4  & 59.8   &  39.9    &    R50GN+WS       &  40.6  & 61.3   &  43.9  & 36.6   &  58.2  &  39.1      \\
\hline
w/o GC     &   X101-64x4d      &  42.1 &63.8 &46.3 & 38.0 &60.6& 40.9    \\
w/ GC     &    X101-64x4d     & 42.8  & 64.5 & 46.8  &  38.4  & 61.0   &  41.1    \\
\hline
\end{tabular}\label{Tmask}\vspace{-5mm}
\end{table}

\subsection{Objection Detection and Segmentation}

Finally, we evaluate GC on object detection and segmentation tasks to show that GC can also be applied to more tasks beyond image classification. The models are pre-trained on ImageNet. The training batch size for object detection and segmentation is usually very small (e.g., 1 or 2) because of the high resolution of input image. Therefore, the BN layer is usually frozen~\cite{he2016deep} and the benefits from BN cannot be enjoyed during training. One alternative is to use GN instead. The models are trained on  COCO $train2017$ dataset (118K images) and evaluated on COCO $val2017$ dataset (40K images)~\cite{lin2014microsoft}. COCO dataset can be used for multiple tasks, including image classification, object detection, semantic segmentation and instance segmentation.

We use the MMDetection~\cite{chen2019mmdetection} toolbox, which contains comprehensive models on object detection and segmentation tasks, as the detection framework. The official implementations and settings are used for all experiments. All the pre-trained models are provided from their official websites, and we fine-tune them on COCO $train2017$ set with 8 GPUs and 2 images per GPU. The optimizers are SGDM and SGDM+GC. The backbone networks include ResNet50 (R50), ResNet101 (R101), ResNeXt101-32x4d (X101-32x4d), ResNeXt101-64x4d (X101-32x4d). The Feature Pyramid Network (FPN)~\cite{lin2017feature} is also used.
The learning rate schedule is $1{X}$ for both Faster R-CNN~\cite{ren2015faster} and Mask R-CNN~\cite{he2017mask}, except R50 with GN and R101 with GN, which use $2{X}$ learning rate schedule.

Tabel \ref{Tfaster} shows the Average Precision (AP) results of Faster R-CNN. We can see that all the
backbone networks trained with GC can achieve a performance gain about $0.3\%\sim0.6\%$ on object detection. Tabel \ref{Tmask} presents the Average Precision  for bounding box ($\text{AP}^b$) and instance segmentation ($\text{AP}^m$). It can be seen that the $\text{AP}^b$ increases by $0.5\%\sim0.9\%$ for object detection task and the $\text{AP}^m$ increases by $0.3\%\sim0.7\%$ for instance segmentation task. Moreover, we find that if  4conv1fc bounding box head, like R50 (4c1f), is used, the performance can increase more by GC. And GC can also boost the performance of GN (see R101GN) and improve the performance of WS (see R50GN+WS). Overall, we can see that GC  boosts the generalization performance of all evaluated models.
This demonstrates that it is a simple yet effective optimization technique, which is general to many tasks beyond image classification.

\section{Conclusions}
How to efficiently and effectively optimize a DNN is one of the key issues in deep learning research. Previous methods such as batch normalization (BN) and weight standardization (WS) mostly operate on network activations or weights to improve DNN training. We proposed a different approach which operates directly on  gradients. Specifically, we removed the mean from the gradient vectors and centralized them to have zero mean. The so-called Gradient Centralization (GC) method demonstrated several desired properties of deep network optimization. We showed that GC actually improves the loss function with a constraint on weight vectors, which regularizes both weight space and output feature space. We also showed that this constrained loss function
 has better  Lipschitzness than the original one so that it has a smoother optimization landscape. Comprehensive experiments were performed and the results demonstrated that GC can be well applied to different tasks with different optimizers and network architectures, improving their training efficiency and generalization performance.

\section*{Appendix}

\subsection*{A1. Proof of Theorem 4.1}
\vspace{-4mm}
\bigskip\noindent\emph{Proof}.
First we show below a simple property of $\mathbf{P}$:
\vspace{-2mm}
$$\mathbf{1}^T\mathbf{P}=\mathbf{1}^T(\mathbf{I}-\mathbf{e}\mathbf{e}^{T})=\mathbf{1}^T-\frac{1}{M}\mathbf{1}^T\mathbf{1}\mathbf{1}^T=\mathbf{0}^T,\vspace{-3mm}$$
where $M$ is the dimension of $\mathbf{e}$.

For each SGD step with GC, we have:
\vspace{-2mm}$$\mathbf{w}^{t+1}=\mathbf{w}^{t}-\alpha^t\mathbf{P}\nabla_{\mathbf{w}^t}\mathcal{L}.\vspace{-3mm}$$
It can be easily derived that: \vspace{-4mm} $$\mathbf{w}^{t}=\mathbf{w}^{0}-\mathbf{P}\sum_{i=0}^{t-1}\alpha^{(i)}\nabla_{\mathbf{w}^{(i)}}\mathcal{L},\vspace{-3mm}$$
where $t$ is the number of iterations.
Then for the  output activations of $\mathbf{x}$ and $\mathbf{x}+\gamma\mathbf{1}$, there is
\vspace{-0mm}
\begin{equation}
\begin{aligned}
({\mathbf{w}^{t}})^T\mathbf{x}-({\mathbf{w}^{t}})^T(\mathbf{x}+\gamma\mathbf{1})&=\gamma \mathbf{1}^T{\mathbf{w}^{t}}\\
&=\gamma \mathbf{1}^T(\mathbf{w}^{0}-\mathbf{P}\sum_{i=0}^{t-1}\alpha^{(i)}\nabla_{\mathbf{w}^{(i)}}\mathcal{L})\\
&=\gamma \mathbf{1}^T\mathbf{w}^{0}-\gamma\mathbf{1}^T\mathbf{P}\sum_{i=0}^{t-1}\alpha^{(i)}\nabla_{\mathbf{w}^{(i)}}\mathcal{L}\\
&=\gamma \mathbf{1}^T\mathbf{w}^{0}.
  \end{aligned}\label{11}
  \vspace{-2mm}
\end{equation}
Therefore,
\begin{equation}
\begin{aligned}
({\mathbf{w}^{t}})^T\mathbf{x}-({\mathbf{w}^{t}})^T(\mathbf{x}+\gamma\mathbf{1})=\gamma\mathbf{1}^T\mathbf{w}^0.
  \end{aligned}\label{12}
  \vspace{-1mm}
\end{equation}

For SGD with momentum, the conclusion is the same, because we can  obtain a  term $\gamma\mathbf{1}^T\mathbf{P}\sum_{i=0}^{t-1}\alpha^{(i)}\mathbf{m}^{i}$ in the third row of Eq.(\ref{11}), where $\mathbf{m}^{i}$ is the momentum in the $i$th iteration, and this term  is also equal to zero.

The proof is completed.
$\blacksquare$

\subsection*{A2. Proof of Theorem 4.2}

\bigskip\noindent\emph{Proof}.
Because $\mathbf{e}$ is a unit vector, there is $\mathbf{e}^T\mathbf{e}=1$.
We can easily prove that:
  \vspace{-2mm}
\begin{equation}
\begin{aligned}
\mathbf{P}^T\mathbf{P}&=(\mathbf{I}-\mathbf{e}\mathbf{e}^{T})^T(\mathbf{I}-\mathbf{e}\mathbf{e}^{T})\\
&=\mathbf{I}-2\mathbf{e}\mathbf{e}^{T}+\mathbf{e}\mathbf{e}^{T}\mathbf{e}\mathbf{e}^{T}\\
&=\mathbf{I}-\mathbf{e}\mathbf{e}^{T}\\
&=\mathbf{P}.
  \end{aligned}
  \vspace{-2mm}
\end{equation}
Then for $\Phi_{GC}(\nabla_{\mathbf{w}}\mathcal{L})$, we have:
 \vspace{-2mm}
\begin{equation}
\begin{aligned}
||\Phi_{GC}(\nabla_{\mathbf{w}}\mathcal{L})||_2^2&=\Phi_{GC}(\nabla_{\mathbf{w}}\mathcal{L})^T\Phi_{GC}(\nabla_{\mathbf{w}}\mathcal{L})\\
&=(\mathbf{P}\nabla_{\mathbf{w}}\mathcal{L})^T(\mathbf{P}\nabla_{\mathbf{w}}\mathcal{L})\\
&=\nabla_{\mathbf{w}}\mathcal{L}^T\mathbf{P}^T\mathbf{P}\nabla_{\mathbf{w}}\mathcal{L}\\
&=\nabla_{\mathbf{w}}\mathcal{L}^T\mathbf{P}\nabla_{\mathbf{w}}\mathcal{L}\\
&=\nabla_{\mathbf{w}}\mathcal{L}^T(\mathbf{I}-\mathbf{e}\mathbf{e}^{T})\nabla_{\mathbf{w}}\mathcal{L}\\
&=\nabla_{\mathbf{w}}\mathcal{L}^T\nabla_{\mathbf{w}}\mathcal{L}- \nabla_{\mathbf{w}}\mathcal{L}^T\mathbf{e}\mathbf{e}^{T}\nabla_{\mathbf{w}}\mathcal{L}\\
&=||\nabla_{\mathbf{w}}\mathcal{L}||_2^2-||\mathbf{e}^{T}\nabla_{\mathbf{w}}\mathcal{L}||_2^2\\
&\leq ||\nabla_{\mathbf{w}}\mathcal{L}||_2^2.
  \end{aligned}
  \vspace{-2mm}
\end{equation}
For  $\nabla_{\mathbf{w}}\Phi_{GC}(\nabla_{\mathbf{w}}\mathcal{L})$, we also have
\begin{equation}
\begin{aligned}
||\nabla\Phi_{GC}(\nabla_{\mathbf{w}}\mathcal{L})||_2^2&=||\mathbf{P}\nabla^2_{\mathbf{w}}\mathcal{L}||_2^2\\
&=\nabla^2_{\mathbf{w}}\mathcal{L}^T\mathbf{P}^T\mathbf{P}\nabla^2_{\mathbf{w}}\mathcal{L}\\
&=\nabla^2_{\mathbf{w}}\mathcal{L}^T\mathbf{P}\nabla^2_{\mathbf{w}}\mathcal{L}\\
&=||\nabla^2_{\mathbf{w}}\mathcal{L}||_2^2-||\mathbf{e}^{T}\nabla^2_{\mathbf{w}}\mathcal{L}||_2^2\\
&\leq ||\nabla^2_{\mathbf{w}}\mathcal{L}||_2^2.
  \end{aligned}
  \vspace{-2mm}
\end{equation}

The proof is completed.
$\blacksquare$

\clearpage
%
%
%
%


\begin{thebibliography}{10}
\providecommand{\url}[1]{\texttt{#1}}
\providecommand{\urlprefix}{URL }
\providecommand{\doi}[1]{https://doi.org/#1}

\bibitem{abadi2016deep}
Abadi, M., Chu, A., Goodfellow, I., McMahan, H.B., Mironov, I., Talwar, K.,
  Zhang, L.: Deep learning with differential privacy. In: Proceedings of the
  2016 ACM SIGSAC Conference on Computer and Communications Security. pp.
  308--318. ACM (2016)

\bibitem{abu2016youtube}
Abu-El-Haija, S., Kothari, N., Lee, J., Natsev, P., Toderici, G., Varadarajan,
  B., Vijayanarasimhan, S.: Youtube-8m: A large-scale video classification
  benchmark. arXiv preprint arXiv:1609.08675  (2016)

\bibitem{Bjorck2018Understanding}
Bjorck, J., Gomes, C., Selman, B., Weinberger, K.Q.: Understanding batch
  normalization pp. 7694--7705 (2018)

\bibitem{bottou1991stochastic}
Bottou, L.: Stochastic gradient learning in neural networks. Proceedings of
  Neuro-N{\i}mes  \textbf{91}(8), ~12 (1991)

\bibitem{bottou2010large}
Bottou, L.: Large-scale machine learning with stochastic gradient descent. In:
  Proceedings of COMPSTAT'2010, pp. 177--186. Springer (2010)

\bibitem{chen2019mmdetection}
Chen, K., Wang, J., Pang, J., Cao, Y., Xiong, Y., Li, X., Sun, S., Feng, W.,
  Liu, Z., Xu, J., et~al.: Mmdetection: Open mmlab detection toolbox and
  benchmark. arXiv preprint arXiv:1906.07155  (2019)

\bibitem{cho2017riemannian}
Cho, M., Lee, J.: Riemannian approach to batch normalization. In: Advances in
  Neural Information Processing Systems. pp. 5225--5235 (2017)

\bibitem{courbariaux2015binaryconnect}
Courbariaux, M., Bengio, Y., David, J.P.: Binaryconnect: Training deep neural
  networks with binary weights during propagations. In: Advances in neural
  information processing systems. pp. 3123--3131 (2015)

\bibitem{courbariaux2016binarized}
Courbariaux, M., Hubara, I., Soudry, D., El-Yaniv, R., Bengio, Y.: Binarized
  neural networks: Training deep neural networks with weights and activations
  constrained to+ 1 or-1. arXiv preprint arXiv:1602.02830  (2016)

\bibitem{duchi2011adaptive}
Duchi, J., Hazan, E., Singer, Y.: Adaptive subgradient methods for online
  learning and stochastic optimization. Journal of Machine Learning Research
  \textbf{12}(Jul),  2121--2159 (2011)

\bibitem{glorot2010understanding}
Glorot, X., Bengio, Y.: Understanding the difficulty of training deep
  feedforward neural networks. In: Proceedings of the thirteenth international
  conference on artificial intelligence and statistics. pp. 249--256 (2010)

\bibitem{gupta2018cnn}
Gupta, H., Jin, K.H., Nguyen, H.Q., McCann, M.T., Unser, M.: Cnn-based
  projected gradient descent for consistent ct image reconstruction. IEEE
  transactions on medical imaging  \textbf{37}(6),  1440--1453 (2018)

\bibitem{he2017mask}
He, K., Gkioxari, G., Doll{\'a}r, P., Girshick, R.: Mask r-cnn. In: Proceedings
  of the IEEE international conference on computer vision. pp. 2961--2969
  (2017)

\bibitem{he2015delving}
He, K., Zhang, X., Ren, S., Sun, J.: Delving deep into rectifiers: Surpassing
  human-level performance on imagenet classification. In: Proceedings of the
  IEEE international conference on computer vision. pp. 1026--1034 (2015)

\bibitem{he2016deep}
He, K., Zhang, X., Ren, S., Sun, J.: Deep residual learning for image
  recognition. In: Proceedings of the IEEE conference on computer vision and
  pattern recognition. pp. 770--778 (2016)

\bibitem{huang2017densely}
Huang, G., Liu, Z., Van Der~Maaten, L., Weinberger, K.Q.: Densely connected
  convolutional networks. In: Proceedings of the IEEE conference on computer
  vision and pattern recognition. pp. 4700--4708 (2017)

\bibitem{huang2017centered}
Huang, L., Liu, X., Liu, Y., Lang, B., Tao, D.: Centered weight normalization
  in accelerating training of deep neural networks. In: Proceedings of the IEEE
  International Conference on Computer Vision. pp. 2803--2811 (2017)

\bibitem{huang2017arbitrary}
Huang, X., Belongie, S.: Arbitrary style transfer in real-time with adaptive
  instance normalization. In: Proceedings of the IEEE International Conference
  on Computer Vision. pp. 1501--1510 (2017)

\bibitem{ioffe2015batch}
Ioffe, S., Szegedy, C.: Batch normalization: Accelerating deep network training
  by reducing internal covariate shift. arXiv preprint arXiv:1502.03167  (2015)

\bibitem{iscen2019label}
Iscen, A., Tolias, G., Avrithis, Y., Chum, O.: Label propagation for deep
  semi-supervised learning. In: Proceedings of the IEEE Conference on Computer
  Vision and Pattern Recognition. pp. 5070--5079 (2019)

\bibitem{karpathy2014large}
Karpathy, A., Toderici, G., Shetty, S., Leung, T., Sukthankar, R., Fei-Fei, L.:
  Large-scale video classification with convolutional neural networks. In:
  Proceedings of the IEEE conference on Computer Vision and Pattern
  Recognition. pp. 1725--1732 (2014)

\bibitem{khosla2011novel}
Khosla, A., Jayadevaprakash, N., Yao, B., Li, F.F.: Novel dataset for fgvc:
  Stanford dogs. In: San Diego: CVPR Workshop on FGVC. vol.~1 (2011)

\bibitem{kim2016accurate}
Kim, J., Kwon~Lee, J., Mu~Lee, K.: Accurate image super-resolution using very
  deep convolutional networks. In: Proceedings of the IEEE conference on
  computer vision and pattern recognition. pp. 1646--1654 (2016)

\bibitem{kingma2014adam}
Kingma, D.P., Ba, J.: Adam: A method for stochastic optimization. arXiv
  preprint arXiv:1412.6980  (2014)

\bibitem{krause20133d}
Krause, J., Stark, M., Deng, J., Fei-Fei, L.: 3d object representations for
  fine-grained categorization. In: Proceedings of the IEEE International
  Conference on Computer Vision Workshops. pp. 554--561 (2013)

\bibitem{krizhevsky2009learning}
Krizhevsky, A., Hinton, G., et~al.: Learning multiple layers of features from
  tiny images. Tech. rep., Citeseer (2009)

\bibitem{krogh1992simple}
Krogh, A., Hertz, J.A.: A simple weight decay can improve generalization. In:
  Advances in neural information processing systems. pp. 950--957 (1992)

\bibitem{larsson2017projected}
Larsson, M., Arnab, A., Kahl, F., Zheng, S., Torr, P.: A projected gradient
  descent method for crf inference allowing end-to-end training of arbitrary
  pairwise potentials. In: International Workshop on Energy Minimization
  Methods in Computer Vision and Pattern Recognition. pp. 564--579. Springer
  (2017)

\bibitem{lei2016layer}
Lei~Ba, J., Kiros, J.R., Hinton, G.E.: Layer normalization. arXiv preprint
  arXiv:1607.06450  (2016)

\bibitem{lin2017feature}
Lin, T.Y., Doll{\'a}r, P., Girshick, R., He, K., Hariharan, B., Belongie, S.:
  Feature pyramid networks for object detection. In: Proceedings of the IEEE
  conference on computer vision and pattern recognition. pp. 2117--2125 (2017)

\bibitem{lin2014microsoft}
Lin, T.Y., Maire, M., Belongie, S., Hays, J., Perona, P., Ramanan, D.,
  Doll{\'a}r, P., Zitnick, C.L.: Microsoft coco: Common objects in context. In:
  European conference on computer vision. pp. 740--755. Springer (2014)

\bibitem{loshchilov2017decoupled}
Loshchilov, I., Hutter, F.: Decoupled weight decay regularization. arXiv
  preprint arXiv:1711.05101  (2017)

\bibitem{luo2018towards}
Luo, P., Wang, X., Shao, W., Peng, Z.: Towards understanding regularization in
  batch normalization  (2018)

\bibitem{maji2013fine}
Maji, S., Rahtu, E., Kannala, J., Blaschko, M., Vedaldi, A.: Fine-grained
  visual classification of aircraft. arXiv preprint arXiv:1306.5151  (2013)

\bibitem{nair2010rectified}
Nair, V., Hinton, G.E.: Rectified linear units improve restricted boltzmann
  machines. In: Proceedings of the 27th international conference on machine
  learning (ICML-10). pp. 807--814 (2010)

\bibitem{pascanu2012understanding}
Pascanu, R., Mikolov, T., Bengio, Y.: Understanding the exploding gradient
  problem. CoRR, abs/1211.5063  \textbf{2} (2012)

\bibitem{pascanu2013difficulty}
Pascanu, R., Mikolov, T., Bengio, Y.: On the difficulty of training recurrent
  neural networks. In: International conference on machine learning. pp.
  1310--1318 (2013)

\bibitem{qian1999momentum}
Qian, N.: On the momentum term in gradient descent learning algorithms. Neural
  networks  \textbf{12}(1),  145--151 (1999)

\bibitem{qiao2019weight}
Qiao, S., Wang, H., Liu, C., Shen, W., Yuille, A.: Weight standardization.
  arXiv preprint arXiv:1903.10520  (2019)

\bibitem{rastegari2016xnor}
Rastegari, M., Ordonez, V., Redmon, J., Farhadi, A.: Xnor-net: Imagenet
  classification using binary convolutional neural networks. In: European
  Conference on Computer Vision. pp. 525--542. Springer (2016)

\bibitem{ravi2016optimization}
Ravi, S., Larochelle, H.: Optimization as a model for few-shot learning  (2016)

\bibitem{ren2015faster}
Ren, S., He, K., Girshick, R., Sun, J.: Faster r-cnn: Towards real-time object
  detection with region proposal networks. In: Advances in neural information
  processing systems. pp. 91--99 (2015)

\bibitem{russakovsky2015imagenet}
Russakovsky, O., Deng, J., Su, H., Krause, J., Satheesh, S., Ma, S., Huang, Z.,
  Karpathy, A., Khosla, A., Bernstein, M., et~al.: Imagenet large scale visual
  recognition challenge. International journal of computer vision
  \textbf{115}(3),  211--252 (2015)

\bibitem{salimans2016weight}
Salimans, T., Kingma, D.P.: Weight normalization: A simple reparameterization
  to accelerate training of deep neural networks. In: Advances in Neural
  Information Processing Systems. pp. 901--909 (2016)

\bibitem{Santurkar2018How}
Santurkar, S., Tsipras, D., Ilyas, A., Madry, A.: How does batch normalization
  help optimization? (no, it is not about internal covariate shift) pp.
  2483--2493 (2018)

\bibitem{simonyan2014very}
Simonyan, K., Zisserman, A.: Very deep convolutional networks for large-scale
  image recognition. arXiv preprint arXiv:1409.1556  (2014)

\bibitem{ulyanov2016instance}
Ulyanov, D., Vedaldi, A., Lempitsky, V.: Instance normalization: The missing
  ingredient for fast stylization. arXiv preprint arXiv:1607.08022  (2016)

\bibitem{vinyals2016matching}
Vinyals, O., Blundell, C., Lillicrap, T., Wierstra, D., et~al.: Matching
  networks for one shot learning. In: Advances in neural information processing
  systems. pp. 3630--3638 (2016)

\bibitem{vorontsov2017orthogonality}
Vorontsov, E., Trabelsi, C., Kadoury, S., Pal, C.: On orthogonality and
  learning recurrent networks with long term dependencies. In: Proceedings of
  the 34th International Conference on Machine Learning-Volume 70. pp.
  3570--3578. JMLR. org (2017)

\bibitem{wah2011caltech}
Wah, C., Branson, S., Welinder, P., Perona, P., Belongie, S.: The caltech-ucsd
  birds-200-2011 dataset  (2011)

\bibitem{wu2018group}
Wu, Y., He, K.: Group normalization. In: Proceedings of the European Conference
  on Computer Vision (ECCV). pp. 3--19 (2018)

\bibitem{xie2017aggregated}
Xie, S., Girshick, R., Doll{\'a}r, P., Tu, Z., He, K.: Aggregated residual
  transformations for deep neural networks. In: Proceedings of the IEEE
  conference on computer vision and pattern recognition. pp. 1492--1500 (2017)

\bibitem{Zhang2016Deep}
Zhang, C., Bengio, S., Hardt, M., Recht, B., Vinyals, O.: Understanding deep
  learning requires rethinking generalization. arXiv preprint arXiv:1611.03530
  (2016)

\bibitem{zhang2018three}
Zhang, G., Wang, C., Xu, B., Grosse, R.: Three mechanisms of weight decay
  regularization. arXiv preprint arXiv:1810.12281  (2018)

\end{thebibliography}
\end{document}